\documentclass[runningheads]{llncs}
\usepackage[T1]{fontenc}
\usepackage{graphicx,verbatim}

\usepackage{booktabs}
\usepackage{amsmath}
\begin{document}
\title{OR-Action: Multi-Role Video Understanding with Fine-Grained Actions}
%

\author{
Felix Tristram\inst{1,2} \and
Ege Özsoy\inst{1,2} \and
Christian Benz\inst{3} \and
Marcel Walch\inst{3} \and
Ghazal Ghazaei\inst{3}\textsuperscript{,*} \and
Nassir Navab\inst{1,2}\textsuperscript{,*}
}


\authorrunning{F. Tristram et al.}

\institute{
Technical University of Munich, Germany\\
\and Munich Center for Machine Learning (MCML)\\
\email{felix.tristram@tum.de}
\and
Carl Zeiss AG, Germany
}

\maketitle
\begingroup
\renewcommand{\thefootnote}{*}
\footnotetext{\scriptsize Equal advising.}
\endgroup              
\begin{abstract}

Fine-grained understanding of operating room (OR) activity could enable workflow-aware assistance, yet remains difficult due to clutter, occlusions, and limited sensing. The prevailing approach to model this environment is scene graphs as an interpretable representation of OR interactions. Converting their frame-wise relational predictions into temporally extended, fine-grained actions however, is challenging without explicit temporal modeling. To enable a principled temporal evaluation of current OR understanding methods, we introduce the first action-centric benchmark built on a publicly available ego-exocentric OR dataset by defining a fine-grained, multi-role action taxonomy and generating dense action segments via distillation from ground-truth scene graph state changes. Experiments on this benchmark show that current scene graph prediction methods struggle to model temporal structure, even when adding explicit modeling through Graph Neural Networks. We therefore introduce a vision-only temporal model that outperforms graph-based methods significantly when using all available egocentric video as input. Building on this model we also introduce a novel multi- to single-view feature alignment strategy that improves single-view performance on multi-role action recognition, mitigating the need for extensive egocentric video capture. Benchmark and code will be released upon acceptance.


\end{abstract}
\section{Introduction}

Operating rooms (ORs) are dense, dynamic, and safety-critical environments in which multiple clinicians coordinate around a patient while interacting with tools, devices, and displays. Reliable perception of this external scene could enable context-aware safety checks, workflow-aware interfaces, automated documentation, and training feedback, aligning with the broader surgical data science vision \cite{maier2017surgical}. 
However, compared to intra-body endoscopic and laparoscopic video analysis—where progress has been driven by large-scale datasets and well-defined tasks such as workflow recognition, instrument presence, segmentation, keypoints, and skill assessment \cite{twinanda2016endonet,hong2020cholecseg8k,heichole,phakir}—external OR understanding remains underexplored due to privacy constraints, heavy occlusions and clutter, and the need to reason jointly about multi-actor interactions over time \cite{srivastav2018mvor,bastian2023know,schmidt2021multi}. Recent work has begun to model the OR as a structured interaction space using semantic scene graphs that encode entities and relations \cite{ozsoy2025mm,ozsoy2025egoexor,ozsoy20224d}, offering an interpretable and queryable abstraction of “who interacts with what.” Yet many clinically meaningful activities, such as tool pick-ups, handovers, equipment positioning, or site preparation, are inherently temporal and defined by state changes rather than single timepoints; mapping per-timepoint relational snapshots to fine-grained actions is therefore brittle, particularly for long-tailed relations, under occlusion, and when action semantics depend on temporal ordering and persistence \cite{ji2020action}. 
In this work, we introduce a dense, fine-grained action benchmark on top of a publicly available OR dataset with scene graph annotations \cite{ozsoy2025egoexor}. Inspired by progress in general video understanding \cite{grauman2024ego,grauman2022ego4d,sener2022assembly101,damen2018scaling}, we define a taxonomy of atomic, role-specific OR actions and generate dense action segments by applying a rule-based mapper to ground-truth scene graphs, using temporal heuristics that convert relation and state transitions into events, yielding 78 action classes spanning 1295 segments over the full dataset. These annotations provide the OR-Action benchmark grounded in structured supervision and enable a principled evaluation of scene-graph-based pipelines as action recognition systems; we show that performance degrades substantially when extrapolating from \emph{predicted} graphs to dense temporal multi-role actions, revealing a gap between relational parsing accuracy and robust temporal action recognition. 

Finally, we show that strong performance is achievable without scene graphs by training a temporal pooling and multi-person classification framework on top of a video foundation model using all available egocentric views. Studying the impact of different viewpoints on action recognition performance of our vision-only model, we find that there is a large variability depending on which roles viewpoint is being observed. Some actions can only be observed by the main actor - e.g. surgeon cutting - while being occluded from other viewpoints. This motivated us to explore a novel feature alignment strategy that aligns video representations from a multi-view teacher to a single view student and enhances the students action recognition performance by making it more aware of other viewpoints.

In total our contributions include (i) the first fine-grained action-centric external OR video understanding benchmark (\emph{OR-Action}), (ii) a principled evaluation of scene graph based methods and our vision only model, and (iii) a novel cross-view feature alignment strategy to improve single-view egocentric action prediction. 
Together, these contributions bring external OR understanding closer to mature intra-body surgical analysis and modern video understanding benchmarks \cite{grauman2024ego,sener2022assembly101,damen2018scaling}. 

\section{Rule-Based Scene-Graph to Action Mapping}
\label{sec:method:mapping}

\begin{figure}[h]
    \centering
    \includegraphics[width=0.7\linewidth]{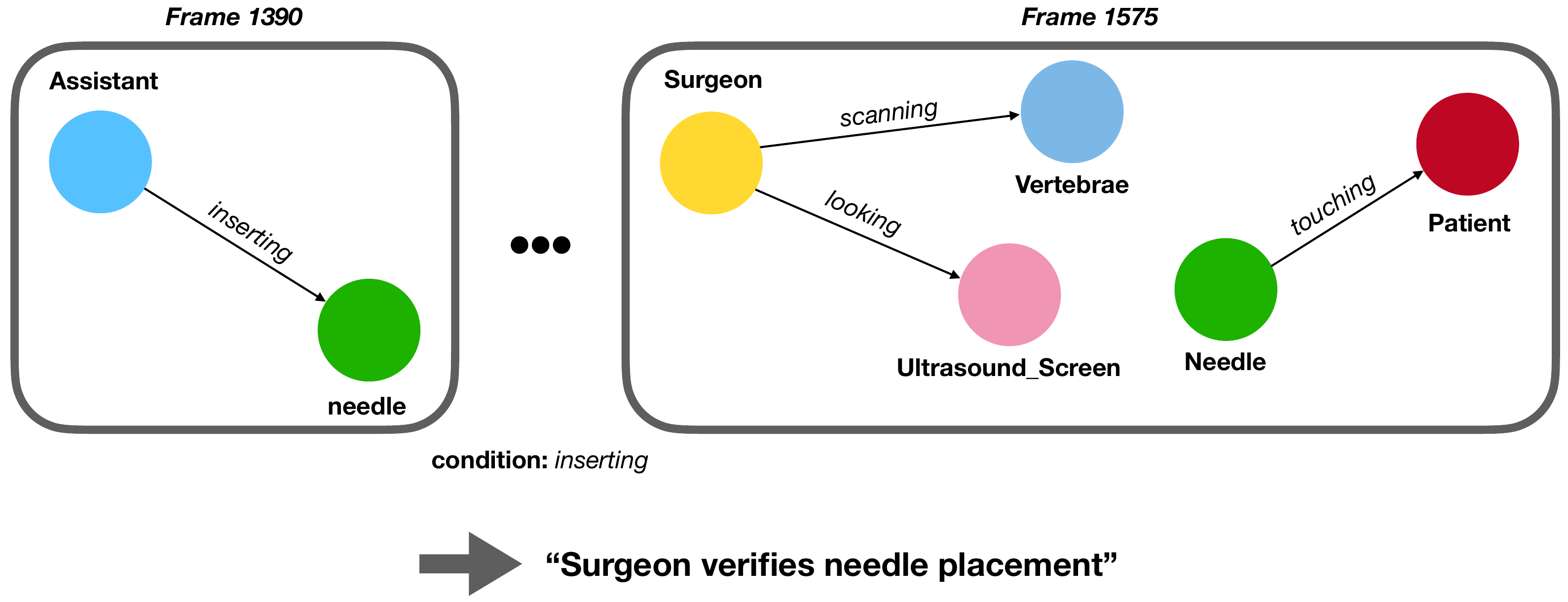}
    \caption{Illustration of "\textit{Surgeon verifies needle placement}" class mapping. From scene graphs this is indistinguishable from \textit{"Surgeon scans for target vertebrae}", so we define the heuristic that this action can only trigger after \textit{``recent''} needle insertion event.}
    \label{fig:mapping_framework}
\end{figure}


We deterministically map per-frame scene graphs from the publicly available EgoExOR dataset \cite{ozsoy2025egoexor} to fine-grained, role-specific action labels and compress them into dense temporal segments. Each original annotated frame is represented by multiple relational triplets (\textit{subject}, \textit{predicate}, \textit{object}). Our rule-based mapper ingests these frame-wise annotations for an entire video and uses them to query a set of predefined rules for an action label. Rules are evaluated independently for every egocentric role in the dataset, assigning either an idle or class label for every role at every time point. 

There are two types of rules: \textbf{State rules}, that trigger whenever boolean conditions over triplets hold and capture persistent activities ((\textit{anaesthetist, looking, health\_monitor}) $\Rightarrow$ ``Anesthetist monitors vitals''). \textbf{Event rules}, that detect change points such as instrument pick-ups, handovers, and returns and convert them into short temporal segments: events are localized to anchor frames and extended into the past and future to reflect movement cues, e.g. the first frame of (\textit{assistant, holding, scalpel}) is the anchor for scalpel pick-up, which then gets extended with past and future frames to reflect the motion of instrument pick-up. 

To improve robustness to annotation noise, we use temporal evidence aggregation - has something happened \textit{``recently''} within a window, or \textit{``ever-before''} in the video. Object handovers are detected from changes in \textit{holding(object)} across roles within a short window; object returns are detected at the end of holding episodes but gated by (\textit{subject,closeTo,instrument-table}) context and brief confirmed absence of the holding relation to suppress brief missing annotations. We prevent repeated pick-ups of the same role--object pair without an intervening return. The rule set is compositional (later decisions may reference earlier outputs) and evaluated in a fixed order; rule conflicts are resolved via priority settings to enforce one label per role per frame. Finally, we merge frame-wise outputs into contiguous segments and apply light smoothing by absorbing very short segments into neighboring actions. This approach yields 1295 action segments over 78 unique classes and almost 100k active and 230k idle frames. Classes are defined from human observation over the entire dataset and refined manually to match behavior between videos. Generated annotations are also visually verified by human annotators for correctness. We re-use the original train/val splits of EgoExOR to maintain compatibility and showcase a visual example with GT and predicted labels in the supplementary video.

\section{Method}
\subsection{Multi-Role Framewise Action Recognition}
\label{sec:method:multirole}

\begin{figure}[h]
    \centering
    \includegraphics[width=1.0\linewidth]{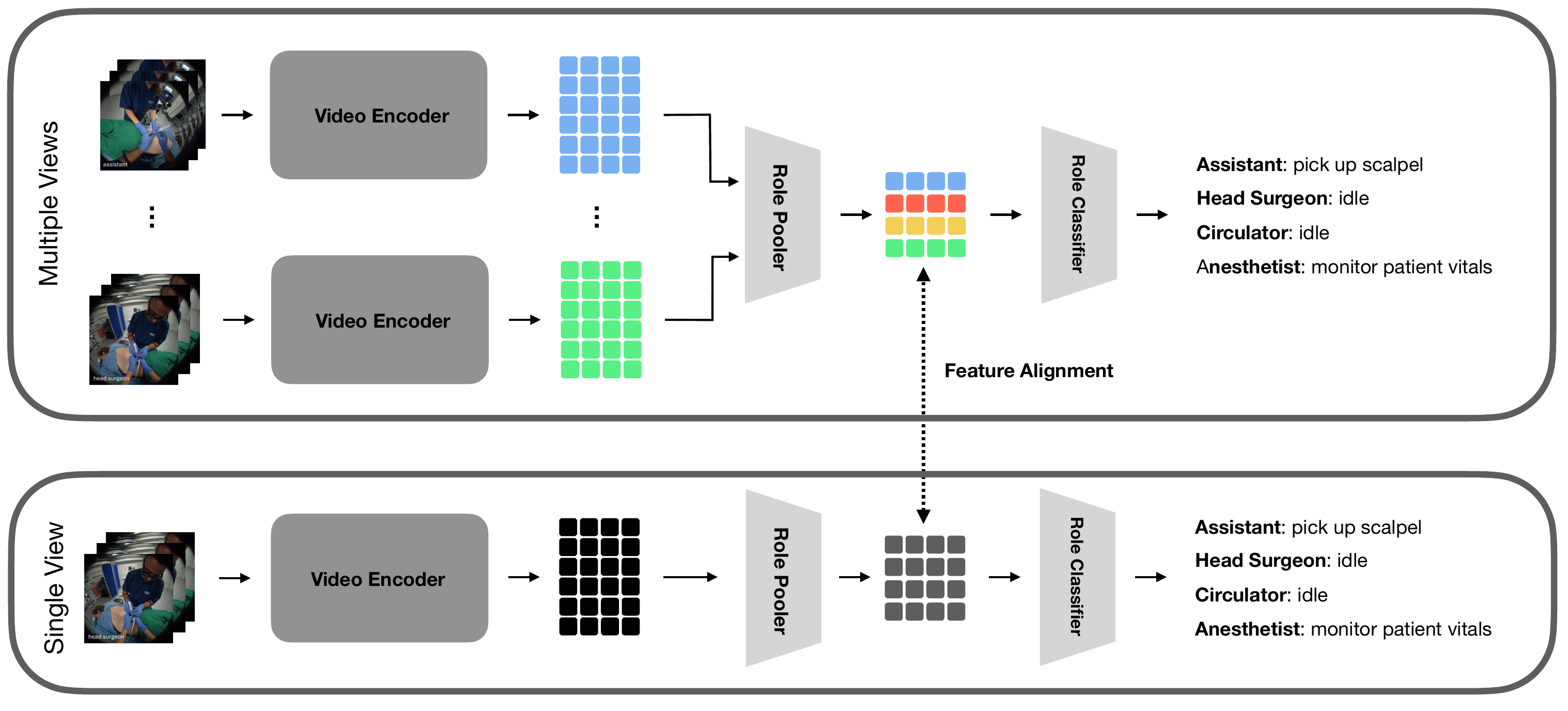}
    \caption{Overview of our vision-only model and cross-view feature alignment strategy}
    \label{fig:method_overview}
\end{figure}

We observe synchronized video clips from a set of camera streams associated with OR \emph{roles} (e.g., different staff members and/or fixed external cameras) as showcased in Figure \ref{fig:method_overview}. 
Let $\mathcal{R}$ denote the target roles for which we predict actions, and $\mathcal{R}_{\mathrm{in}}$ the set of available input streams for a given clip.
For each target role $r \in \mathcal{R}$ and frame index $t \in \{1,\dots,T\}$, we are given (i) a fine-grained action label $y^{(r)}_t \in \{1,\dots,K\}$ and (ii) a binary activity label $a^{(r)}_t \in \{0,1\}$ indicating whether the role is active or idle.

Each input stream $r \in \mathcal{R}_{\mathrm{in}}$ provides a clip $\mathbf{x}^{(r)}_{1:T}$.
Spatiotemporal tokens are extracted with a frozen video foundation model $E$:
\begin{equation}
    \mathbf{z}^{(r)} = E\!\left(\mathbf{x}^{(r)}_{1:T}\right),
\end{equation}
where $\mathbf{z}^{(r)}$ is a sequence of token embeddings.
VJEPA2 \cite{vjepa2} is used as encoder due to its impressive performance representing general motion and semantic cues.

\textbf{Token Compression.}
The encoder produces a long token sequence from which we extract the most informative representations using an attentive Role Pooler $\mathrm{RP}_{Q}(\cdot)$: a set of $Q$ learnable query tokens that attend to the input token set via cross-attention.
For each role we obtain a compact sequence
\begin{equation}
    \mathbf{u}^{(r)} = \mathrm{RP}_{Q_f}\!\left(\mathbf{z}^{(r)}\right),
\end{equation}
and interpret $\mathbf{u}^{(r)}$ as \emph{temporal tokens}. We select $Q$ to obtain exactly $T$ tokens per role, yielding per-frame embeddings $\mathbf{u}^{(r)}_t$.

\textbf{Framewise cross-role fusion.}
To exploit interactions, shared context and mitigate occlusions, we fuse pooled representations between all observed input streams.
For a fixed $t$, we stack available role embeddings $\{\mathbf{u}^{(r)}_t\}_{r \in \mathcal{R}_{\mathrm{in}}}$ and apply a second attentive pooler with $|\mathcal{R}|$ queries as a Role Classifier $RC$, producing one fused embedding per \emph{target} role:
\begin{equation}
    \mathbf{h}_t = \mathrm{RC}_{|\mathcal{R}|}\!\left(\{\mathbf{u}^{(r)}_t\}_{r \in \mathcal{R}_{\mathrm{in}}}\right),
\end{equation}
where $\mathbf{h}_t = \left[\mathbf{h}^{(1)}_t, \dots, \mathbf{h}^{(|\mathcal{R}|)}_t\right]$ and $\mathbf{h}^{(r)}_t$ is the role-conditioned representation used for prediction.
Missing streams are handled by zeroing their tokens and masking losses.

\textbf{Action and activity heads.}
We predict per-frame logits using linear heads:
\begin{equation}
    \boldsymbol{\ell}^{(r)}_t = W_{\mathrm{act}}\, \mathbf{h}^{(r)}_t, \qquad
    \boldsymbol{s}^{(r)}_t = W_{\mathrm{aux}}\, \mathbf{h}^{(r)}_t,
\end{equation}
where $\boldsymbol{\ell}^{(r)}_t$ is a $K$-way action logit vector and $\boldsymbol{s}^{(r)}_t$ is a 2-way activity logit vector.

We use a masked multi-task cross-entropy over roles and frames,
\begin{equation}
    \mathcal{L}_{\mathrm{frame}} = \sum_{r \in \mathcal{R}} \sum_{t=1}^{T} m^{(r)}_t \Big(
    \mathrm{CE}(\mathrm{softmax}(\boldsymbol{\ell}^{(r)}_t), y^{(r)}_t)
    + \mathrm{CE}(\mathrm{softmax}(\boldsymbol{s}^{(r)}_t), a^{(r)}_t) \Big),
\end{equation}
where $m^{(r)}_t$ masks roles that are not available in a procedure.
We choose separate action and activity heads due to large class imbalance between active and idle frames. 

\subsection{Multi- to Single-View Feature Alignment }
\label{sec:method:alignment}

In realistic OR deployments, we may only observe a single wearable stream (or a subset of streams), yet we still want to predict multi-role actions.
We therefore experiment with transferring multi-view context from a \emph{multi-role teacher} to a \emph{single-view student}.

\textbf{Teacher representation.}
We first train the multi-role model from Sec.~\ref{sec:method:multirole} using all available streams, and then freeze it.
Given a clip with streams $\mathcal{R}_{\mathrm{in}}$ (typically $\mathcal{R}_{\mathrm{in}}=\mathcal{R}$), the teacher produces a role-major sequence by pooling each role independently and concatenating:
\begin{equation}
    \mathbf{t} = \big[\, \mathrm{RP}^{T}_{Q_T}(\mathbf{z}^{(r)}) \,\big]_{r \in \mathcal{R}_{\mathrm{in}}}.
\end{equation}
This sequence encodes OR context distributed across roles (what other staff members are doing, shared tool state, etc.).

\textbf{Single-view student and alignment loss.}
The student observes only a fixed input role $r_0$ and produces a sequence of the same length, $\mathbf{s}=\mathrm{RP}^{S}_{Q_S}(\mathbf{z}^{(r_0)})$, where $Q_S$ is chosen so that $\mathbf{s}$ matches the teacher sequence length $(R_{in}\cdot Q)$.
We align tokens using a per-token $\ell_1$ loss:
\begin{equation}
    \mathcal{L}_{\mathrm{align}}
    = \frac{1}{|\Omega|}
    \sum_{i \in \Omega} \left\| \hat{\mathbf{t}}_{i} - \hat{\mathbf{s}}_{i} \right\|_{1},
\end{equation}
where $\Omega$ indexes tokens that correspond to available input streams and $\hat{\cdot}$ denotes $\ell_2$ normalization.
Intuitively, the student is encouraged to extract more contextual cues from the dense output $\mathbf{z}^{(r_0)}$ that would otherwise be missed using $\mathcal{L}_{\mathrm{frame}}$ alone.

\textbf{Joint supervised training.}
To ensure the aligned representation remains predictive of actions, we optimize both alignment and frame-based losses simultaneously.
We reshape the student sequence into role-by-time tokens and apply the same per-frame cross-role fusion and action/activity heads as in Sec.~\ref{sec:method:multirole}, yielding $\boldsymbol{\ell}^{(r)}_t$ and $\boldsymbol{s}^{(r)}_t$ from the single input stream.
The final objective is a weighted combination of alignment and supervised losses:
\begin{equation}
    \mathcal{L} = \lambda_{\mathrm{align}}\, \mathcal{L}_{\mathrm{align}} + \lambda_{\mathrm{sup}}\, \mathcal{L}_{\mathrm{frame}}.
\end{equation}

\section{Experiments}

\begin{figure}[h]
    \centering
    \includegraphics[width=1.0\linewidth]{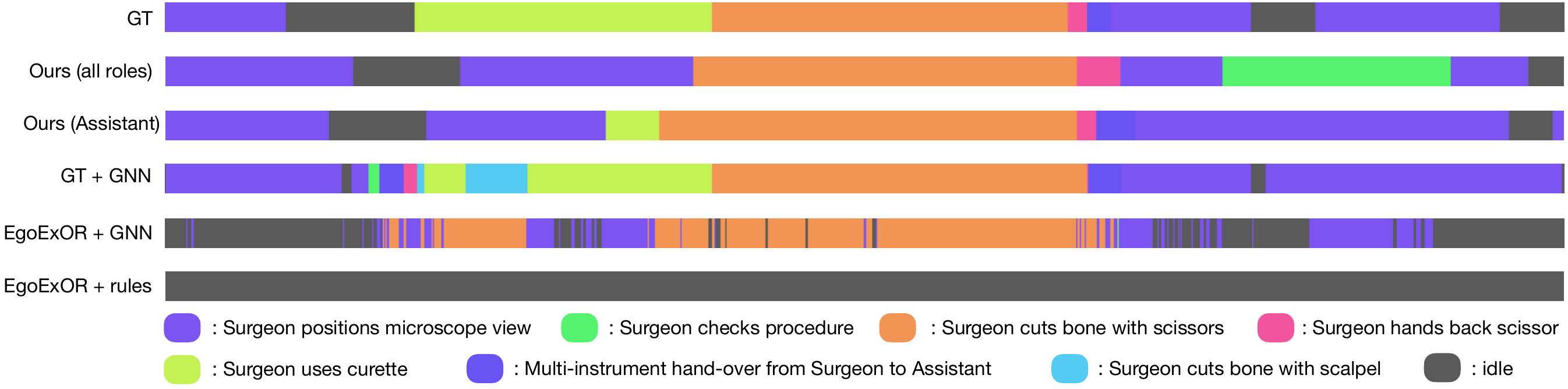}
    \caption{Qualitatively example on (\textit{MISS/3/take/1}) from validation set.}
    \label{fig:qualitatives}
\end{figure}

\textbf{Implementation Details.}
We sample $T=64$ frames at 4 fps (16s windows) to align with VJEPA2's training \cite{vjepa2}, setting $Q=64$ pooling tokens per role. For alignment we set $\lambda_{align}=10.0, \lambda_{sup}=0.1$ to roughly match loss magnitudes. The OR-Action benchmark comprises $K=78$ classes. Due to class imbalance—\textit{idle} classes account for 69\% of frames—we use separate activity and action prediction heads. The GNN baseline initializes nodes from (\textit{subject, object}) with stable embeddings across timepoints, while also connecting node IDs to adjacent frames $[t-1; t+1]$. Scene graph predictions are obtained using the publicly available model checkpoint of EgoExOR~\cite{ozsoy2025egoexor}. At inference, all learned models run in a sliding window fashion: the activity head gates idle frames, after which the action head classifies active frames. Windows overlap by 50\%, with predictions merged via a Hanning-weighted sum to reduce boundary artifacts.

\textbf{Metrics.} 
We report traditional temporal action-segmentation metrics: Frame Accuracy (\textbf{Acc} \textit{w/idle}) (percentage of correctly classified frames, including idle classes); Frame Accuracy without idle (\textbf{Acc}) for evaluation without class imbalance; \textbf{Edit Score} (normalized edit distance on the predicted action sequence, penalizing ordering errors and over/under-segmentation); and Segmental F1 (\textbf{F1}@\textit{k}) (segment-level F1 at overlap thresholds of 10\%, 25\%, and 50\%, requiring both class and boundary agreement). Higher is better for all metrics.

\begin{table}[t]
    \centering
    \setlength{\tabcolsep}{3pt}
    \caption{Benchmarking OR temporal action understanding. Metrics are computed without idle class. GT + GNN is evaluated on GT scene graphs as an upper bound, whereas EgoExOR + GNN is using  predicted scene graphs.}
    \label{tab:val_action_metrics}
    \begin{tabular}{lccccc}
        \toprule
        \textbf{Model} & \textbf{Acc} & \textbf{Edit Score} & \textbf{F1}@(\textit{0.1} & \textit{0.25} & \textit{0.5}) \\
        \midrule
        \multicolumn{6}{l}{\textit{Validation}} \\
        \midrule
        GT + GNN          & 65.19 & 48.33 &  32.08 &  30.42 &  25.42 \\
        \midrule
        EgoExOR + rule mapping      & 16.54 & \textit{38.24} & 21.21 & 16.16 & 13.13 \\
        EgoExOR + GNN    & 36.29 & 30.01          &  3.56 &  2.43 &  1.13 \\
        Ours (all roles)         & \textbf{61.65} & \textbf{38.24} & \textbf{39.90} & \textbf{35.47} & \textbf{25.12} \\
        \midrule
        \multicolumn{6}{l}{\textit{Test}} \\
        \midrule
        GT + GNN         & 69.54 & 48.63 &  39.34 &  36.53 &  29.98 \\
        \midrule
        EgoExOR + rule mapping      & 17.37 & 38.28 & 21.59 & 20.45 & 18.18 \\
        EgoExOR + GNN    & 43.79 & 37.29 &  4.20 &  3.00 &  1.80 \\
        Ours (all roles)         & \textbf{67.50} & \textbf{45.47} & \textbf{40.64} & \textbf{37.88} & \textbf{29.10} \\
        \bottomrule
    \end{tabular}
\end{table}

\textbf{Discussion of Results.}
Table \ref{tab:val_action_metrics} summarizes the main findings of the OR-Action benchmark, with SG denoting scene graph. As a simple baseline, we apply the rule-based mapper from Sec. \ref{sec:method:mapping} to EgoExOR \emph{predicted} SGs. Because these rules are brittle and tuned to ground-truth annotations, we also introduce EgoExOR+GNN, a learned baseline trained on ground-truth SGs but evaluated on EgoExOR \emph{predictions}. The rule-based mapper performs very poorly, underscoring both the brittleness of the rules and the gap between predicted and ground-truth SGs. EgoExOR+GNN improves on this baseline, but still fails to produce coherent temporal segments, as shown by its very low F1-scores and the temporal jitter in Figure \ref{fig:qualitatives}.

To assess scene graphs as a representation independently, we also report an upper bound with GT+GNN, trained \emph{and} evaluated on ground-truth SGs. Its strong performance suggests that SGs can be a useful representation for temporal actions, and that the drop in the predictive setting is mainly due to weak temporal modeling and class imbalance in SG training.

Our vision-only model reaches accuracy and Edit Score close to the oracle GT+GNN, while outperforming it in F1-score at all but the strictest threshold. By explicitly modeling temporal structure and leveraging strong video foundation model features, it captures even brief actions such as the scissor handover in Figure \ref{fig:qualitatives}. Its main limitation remains visually similar actions, such as \emph{surgeon positions microscope} and \emph{surgeon checks procedure}, which are hard to distinguish from vision alone.

\begin{table}[t]
    \centering
    \setlength{\tabcolsep}{3pt}
    \caption{Ablations on viewpoints, feature alignment strategy and split prediction head design. All experiments use our vision-only model and report test-set results. 
    }
    \label{tab:alignement_results}
    \begin{tabular}{lcccccc}
        \toprule
        \textbf{Input Roles} & \textbf{Acc}\textit{(w/idle)} & \textbf{Acc} & \textbf{Edit} & \textbf{F1}@(\textit{0.1} & \textit{0.25} & \textit{0.5}) \\
        \midrule
        all roles (\textit{ego})       & 62.13 & 67.50 & 45.47 &  40.64 &  37.88 &  29.10 \\
        all roles (\textit{no activity head})       & 11.33 & 38.57 & 17.67 &  24.92 &  19.27 &  10.63 \\
        only exo views       & 50.05 & 44.94 & 30.99 &  26.60 &  23.50 &  13.30 \\
        \midrule
        Surgeon     & 57.44 & 57.81 & 30.82 & \textbf{32.46} & \textbf{27.70} & 16.88 \\
        Surgeon\textit{(align)} & \textbf{62.11}     & \textbf{58.16} & \textbf{42.93} & 31.29 & 26.64 & \textbf{19.45} \\
        \midrule
        Assistant      & 60.39 & 64.79 & 40.14 & 38.84 & 36.16 & 24.11 \\
        Assistant\textit{(align)}     & \textbf{67.89} & \textbf{64.90} & \textbf{44.84} & \textbf{41.03} & \textbf{36.36} & \textbf{26.10} \\
        \midrule
        Circulator      & \textbf{36.02} & \textbf{40.28} & 24.12 & 22.58 & \textbf{17.74} & \textbf{9.68} \\
        Circulator\textit{(align)}     & 35.49 & 38.31 & \textbf{30.61} & \textbf{23.53} & 15.68 & 8.62 \\
        \bottomrule
    \end{tabular}
\end{table}

Table \ref{tab:alignement_results} summarizes our experiments with the feature alignment strategy from Sec. \ref{sec:method:alignment}, together with ablations on split prediction heads and ego- versus exocentric inputs. Using a single prediction head (\textit{no activity head}) yields poor performance because the \textit{idle} class dominates the data, accounting for 69\% of all labeled frames. Relying only on exocentric inputs (e.g., ceiling-mounted cameras) also leaves a large gap to egocentric views, as these cameras are fixed, far from the active regions, and often occluded by staff or equipment, making fine movements and small instruments difficult to capture. The strong gain from egocentric inputs highlights the importance of close-up, human-viewpoint observations for reliable OR action understanding.

But not all egocentric viewpoints are equally informative. When using only the input stream from the Circulator, which is mostly inactive and observing the procedure from far away, performance is generally weak and, except for Edit Score, does not benefit from cross-view feature alignment. This is not unexpected, since the feature alignment relies on picking up contextual cues from its single view input, which are simply not observable in this role. 

The opposite is true for the head surgeon and assistants viewpoints. Here we can observe large gains when using the feature alignment strategy, especially in Acc(\textit{w/idle}) and Edit Score, indicating improved awareness of all roles idle/active states and superior action ordering, respectively.
\section{Conclusion}

We present OR-Action, the first multi-role OR benchmark for temporal action understanding, heuristically derived from structured scene graph annotations in the publicly available EgoExOR \cite{ozsoy2025egoexor} dataset. Experiments show that current scene graph prediction models, while strong at scene graph prediction itself, struggle to represent temporal actions. Motivated by this, we introduce a vision-only model that clearly outperforms predicted scene graphs, as well as a novel multi- to single-view egocentric feature alignment strategy that reduces the need for extensive multi-person video capture. We hope OR-Action will accelerate research in external OR understanding and support the development of temporally aware methods in this domain. 

\bibliographystyle{splncs04}
\bibliography{bib}
\end{document}